\documentclass[10pt,twocolumn,letterpaper]{article}

\usepackage{cvpr}              

\usepackage{graphicx}
\usepackage{amsmath}
\usepackage{amssymb}
\usepackage{booktabs}
\usepackage{bbm}

\usepackage[pagebackref,breaklinks,colorlinks]{hyperref}

\usepackage[capitalize]{cleveref}
\crefname{section}{Sec.}{Secs.}
\Crefname{section}{Section}{Sections}
\Crefname{table}{Table}{Tables}
\crefname{table}{Tab.}{Tabs.}

\usepackage{multirow}

\def\cvprPaperID{6199} 
\def\confName{CVPR}
\def\confYear{2022}

\begin{document}

\twocolumn[{
\begin{@twocolumnfalse}

\title{Non-parametric Depth Distribution Modelling based Depth Inference for Multi-view Stereo}

\author{Jiayu Yang$^{1,2^*}$, \;\; Jose M. Alvarez$^2$,\;\; Miaomiao Liu$^1$\\
$^1$Australian National University, $^2$NVIDIA\\
{\tt\small \{jiayu.yang, miaomiao.liu\}@anu.edu.au,}\;\;{\tt\small josea@nvidia.com}
}
\maketitle

\thispagestyle{empty}

\begin{center}
\setlength\tabcolsep{1pt}
\vspace{-0.55cm}
\small
\begin{tabular}{cccccc}
\multirow{3}{*}[1.66cm]{\includegraphics[width=0.32\linewidth]{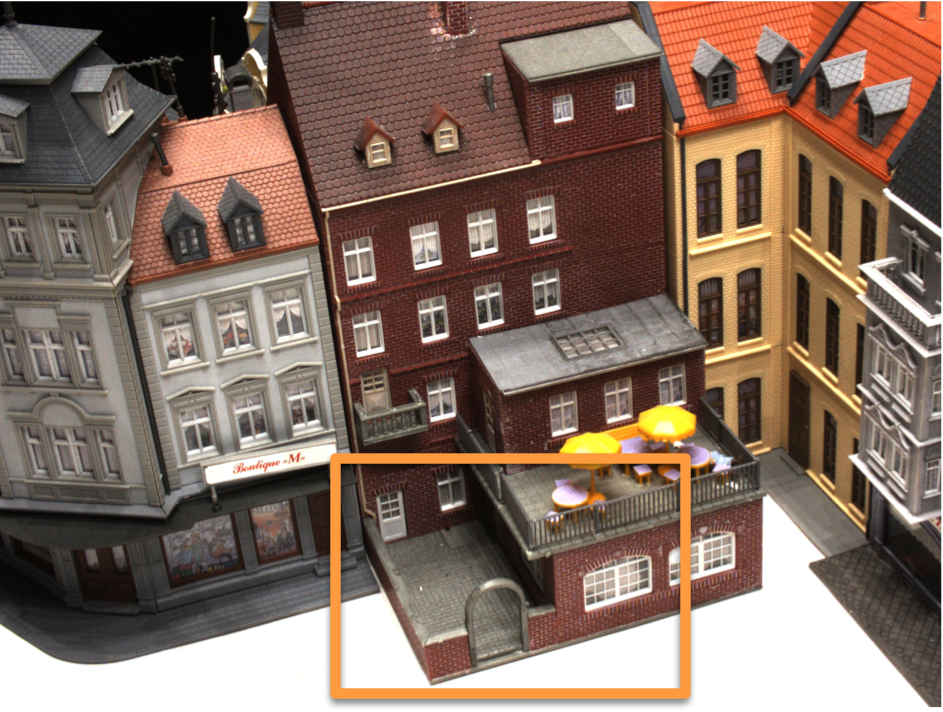}} 
&  \includegraphics[width=0.16\linewidth]{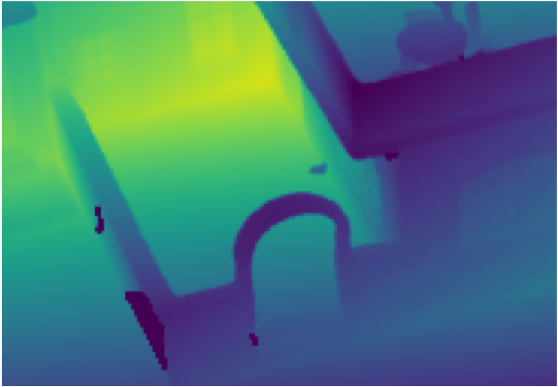}
&  \includegraphics[width=0.16\linewidth]{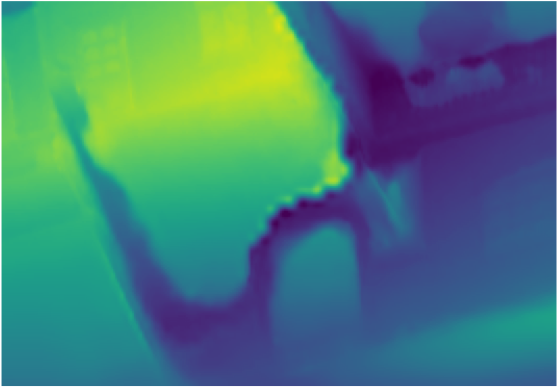}
&  \includegraphics[width=0.16\linewidth]{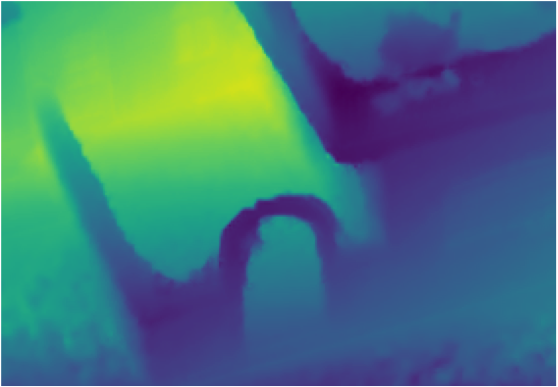}
&  \includegraphics[width=0.16\linewidth]{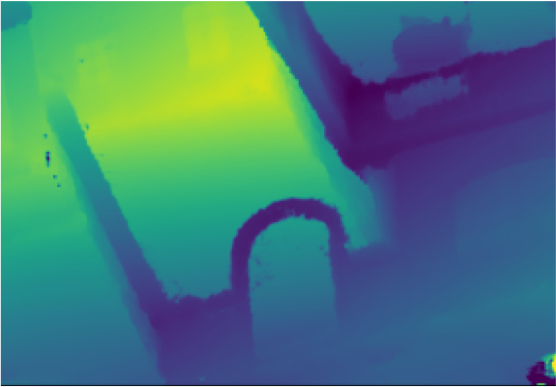}
&  \includegraphics[width=0.039\linewidth]{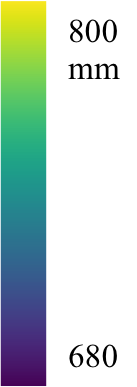} \\
&  Ref. Depth
&  Unimodal baseline
&  PatchmatchNet\cite{wang2020patchmatchnet}
&  Ours
&                   \\    
&  \includegraphics[width=0.16\linewidth]{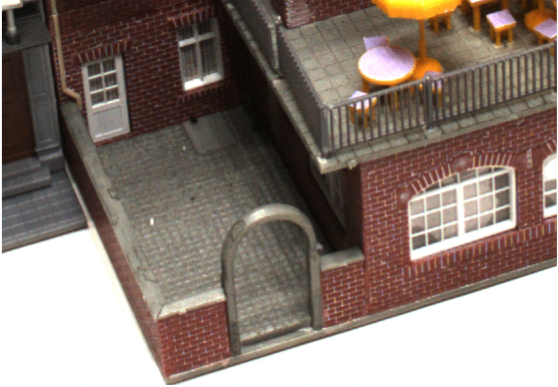}
&  \includegraphics[width=0.16\linewidth]{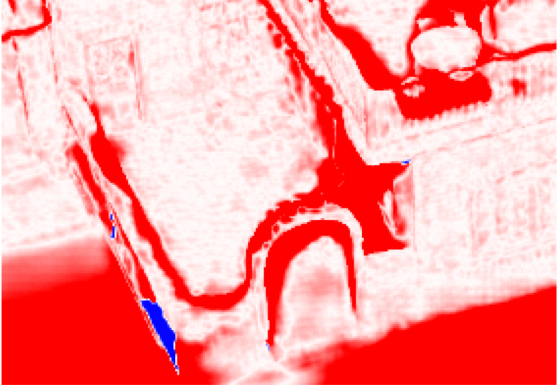}
&  \includegraphics[width=0.16\linewidth]{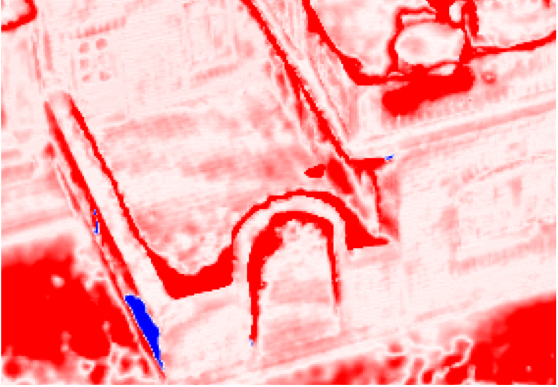}
&  \includegraphics[width=0.16\linewidth]{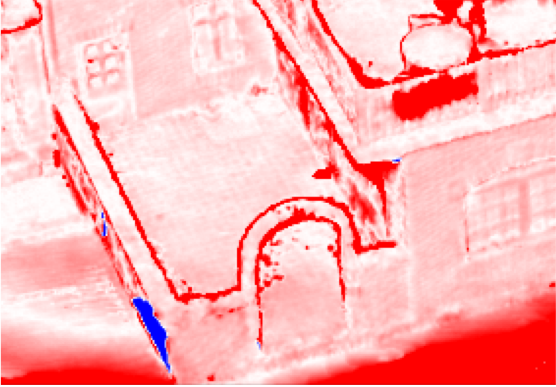}
&  \includegraphics[width=0.039\linewidth]{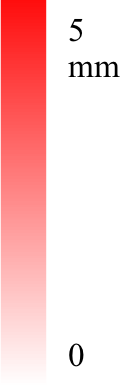} \\
Ref. Image
&  Ref. Image
&  Error of baseline
&  Error of \cite{wang2020patchmatchnet}
&  Error of ours
&                   \\             
\end{tabular}
\end{center}
\vspace{-0.3cm}
\captionof{figure}{Performance analysis of cascade MVS methods on boundary regions. Compared with the~\emph{unimodal~baseline} model, and the state-of-the-art patchmatch based approach~\emph{PatchmatchNet}~\cite{wang2020patchmatchnet}, our method based on non-parametric depth distribution modeling can achieve more accurate depth estimation results on the boundary regions with abrupt depth changes.
}\label{fig:startfig}
\vspace{0.4cm}
\end{@twocolumnfalse}
}]
\begin{abstract}
\vspace{-0.3cm}
Recent cost volume pyramid based deep neural networks have
unlocked the potential of efficiently leveraging high-resolution 
images for depth inference from multi-view stereo. In general,
those approaches assume that the depth of each pixel follows a
unimodal distribution. Boundary pixels usually follow a multi-modal distribution as they represent different depths; Therefore, the assumption results in an erroneous depth prediction at the coarser level of the cost volume pyramid and can not be corrected in the refinement levels leading to wrong depth predictions. In contrast, we propose constructing the cost volume by non-parametric depth distribution modeling to handle pixels with unimodal and multi-modal distributions. Our approach outputs multiple depth hypotheses at the coarser level to avoid errors in the early stage. As we perform local search around these multiple hypotheses in subsequent levels, our approach does not maintain the rigid depth spatial ordering and, therefore, we introduce a sparse cost aggregation network to derive information within each volume. We evaluate our approach extensively on two benchmark datasets: DTU and Tanks \& Temples. Our experimental results show that our model outperforms existing methods by a large margin and achieves superior performance on boundary regions. 

Code is available at \url{https://github.com/NVlabs/NP-CVP-MVSNet}

\end{abstract}

\section{Introduction}
Multi-view stereo (MVS) aims to infer the 3D structure, represented by a depth map, of a scene from a set of images captured by a camera from multiple viewpoints. It is a fundamental problem for the computer vision community and has been studied extensively for decades~\cite{seitz2006comparison}. \let\thefootnote\relax\footnote{$^*$The work is done during an internship at NVIDIA}

Recently, learning-based methods using a cost volume pyramid~\cite{yang2020cvp,cheng2020ucsnet,gu2020cascade} have emerged as the preferred technique to leverage high-resolution images leading to superior performance compared to other learning-based approaches~\cite{yao2018mvsnet,yao2019recurrent}. These approaches usually perform a local depth search around an initial depth estimate on a cost volume built at the coarser level. Then, assuming each pixel follows a unimodal distribution, they estimate the depth for each pixel as the expectation of approximated continuous depth distributions by a few depth samples within a predefined range~\cite{yang2020cvp,cheng2020ucsnet,gu2020cascade}. While these approaches have achieved promising results, they tend to miss small objects and 
boundary regions with abrupt depth changes, where the unimodal distribution assumption does not hold. 

As shown in Fig.~\ref{fig:startfig}, the multi-scale depth estimation frameworks with unimodal distribution assumption performs badly on boundary pixels. It is mainly because existing multi-scale approaches make an early decision at the coarse level as the depth in this level is represented as a single value based on the unimodal distribution assumption.  If the estimated coarse depth is far from the actual depth, the error will be propagated to refinement levels and can not be corrected via local depth search, resulting in incorrect depth prediction.

In this work, we address this problem by explicitly modeling the depth of each pixel at different resolutions using a multi-modal distribution. In particular, in contrast to methods using a mixture density distribution~\cite{tosi2021smd}, we use a non-parametric distribution to learn the probability of each depth hypothesis along the 3D visual ray. Our approach provides additional flexibility compared to parametric approaches, especially in the coarse-to-fine structure. We then guide the learning process using the depth distribution within its corresponding depth patch at the highest resolution. Given the learned distribution, we build the cost volume for the next level by branching the depth hypotheses using the top K probabilities. Our approach achieves a better ground truth depth covering ratio than the existing approaches; however, it loses the relative spatial relationship due to the pixel-wise depth branching processing. To aggregate the information in our new
cost volume structure, we propose a sparse cost aggregation network to preserve the relative spatial relationship. Extensive experiments on several benchmark datasets demonstrate that our approach achieves superior performance, especially on boundary regions. On the DTU dataset, our approach outperforms the current state-of-the-art multi-scale patchmatch based approach PatchmatchNet~\cite{wang2020patchmatchnet}, yielding up to a $32\%$ lower error on boundary regions.

In summary, the contributions in this paper are as follows.
\begin{itemize}
    \item We propose a non-parametric depth probability distribution modeling, allowing us to handle pixels with unimodal and multimodal distributions.
    \item We build a cost volume pyramid by branching the depth samples based on the modeled pixel-wise depth probability distribution.
    \item We apply a sparse cost aggregation network to process each cost volume to maintain rigid geometric spatial relation in the cost volume and avoid spatial ambiguity.
    \item Our approach outperforms previous approaches on boundary areas and becomes the new state of the art on the DTU dataset. 
    \end{itemize}

\section{Related works}
\noindent \textbf{Deep Learning-based MVS} methods adopt deep CNNs to infer the depth map for each view followed by a separate multiple-view fusion process for building 3D models.~These methods allow the network to extract discriminative features encoding global and local information of a scene to obtain robust feature matching for MVS. In particular, Yao \etal propose MVSNet~\cite{yao2018mvsnet} to infer a depth map for each view. An essential step in~\cite{yao2018mvsnet} is to build a cost volume based on a plane sweep process followed by multi-scale 3D CNNs for regularization.~While effective in depth inference accuracy, its memory requirement is cubic to the image resolution. To allow handling high resolution images, several methods have been proposed to reduce the memory requirement. Recurrent methods\cite{yao2019recurrent,yan2020d2hc,wei2021aarmvsnet} build cost volume and aggregate matching cost in a recursive manner using recurrent networks such as GRU\cite{yao2019recurrent} or LSTM\cite{yan2020d2hc,wei2021aarmvsnet}. However, recurrent methods usually consume a longer run-time trading for reduced space requirement. 

Another line of research exploits multi-scale framework for coarse-to-fine depth estimation\cite{chen2019point,yang2020cvp,gu2020cascade,cheng2020ucsnet,xu2020pvsnet,wang2020patchmatchnet}. Multi-scale methods build a coarse cost volume in lower resolution to estimate a coarse depth map followed by building partial cost volumes in higher resolution for depth refinement. Although multi-scale methods have achieved promising results in both efficiency and accuracy, two major problems remain unsolved: Early decision on coarser levels and spatial ambiguity of partial cost volumes. In this paper, we address the first problem by a non-parametric depth distribution modelling paired with a novel multi-scale depth estimation framework. We address the second problem by a novel sparse cost volume formulation and a sparse cost aggregation network that retain rigid spatial relation.

\noindent \textbf{Multi-modal disparity distribution modeling}.
Multi-modal disparity probability distribution modeling has been studied
recently in tackling the stereo matching problem. In~\cite{tosi2021smd}, 
Tosi \etal show that the unimodal probability distribution
assumption for each pixel leads to inaccurate disparity estimations when
the actual disparity follows a multi-modal distribution. They thus propose 
to model the continuous disparity as a parametric bi-modal distribution.
Their proposed approach further improves the disparity estimation 
performance at the high resolution especially for boundary pixels. 
While the proposed approach can tackle the multi-modal disparity/depth
modelling, it is not trivial to be directly extended to the cost volume
pyramid structure to avoid resulting in erroneous depth estimation at the
coarse level.

UASNet~\cite{mao2021uasnet} proposed to build the cascaded
cost volume via uncertainty guided depth range estimation and depth sampling.
Their goal is to build a cost volume with depth sampling covers the
multi-modal depth distribution. In particular, they output 4 disparity maps at
each stage. In contrast, we do not output depth estimate at the early stage 
instead of building cost volume based on depth samples with high probability.

\section{Method}
\begin{figure*}
	\begin{center}
    \includegraphics[width=0.85\linewidth]{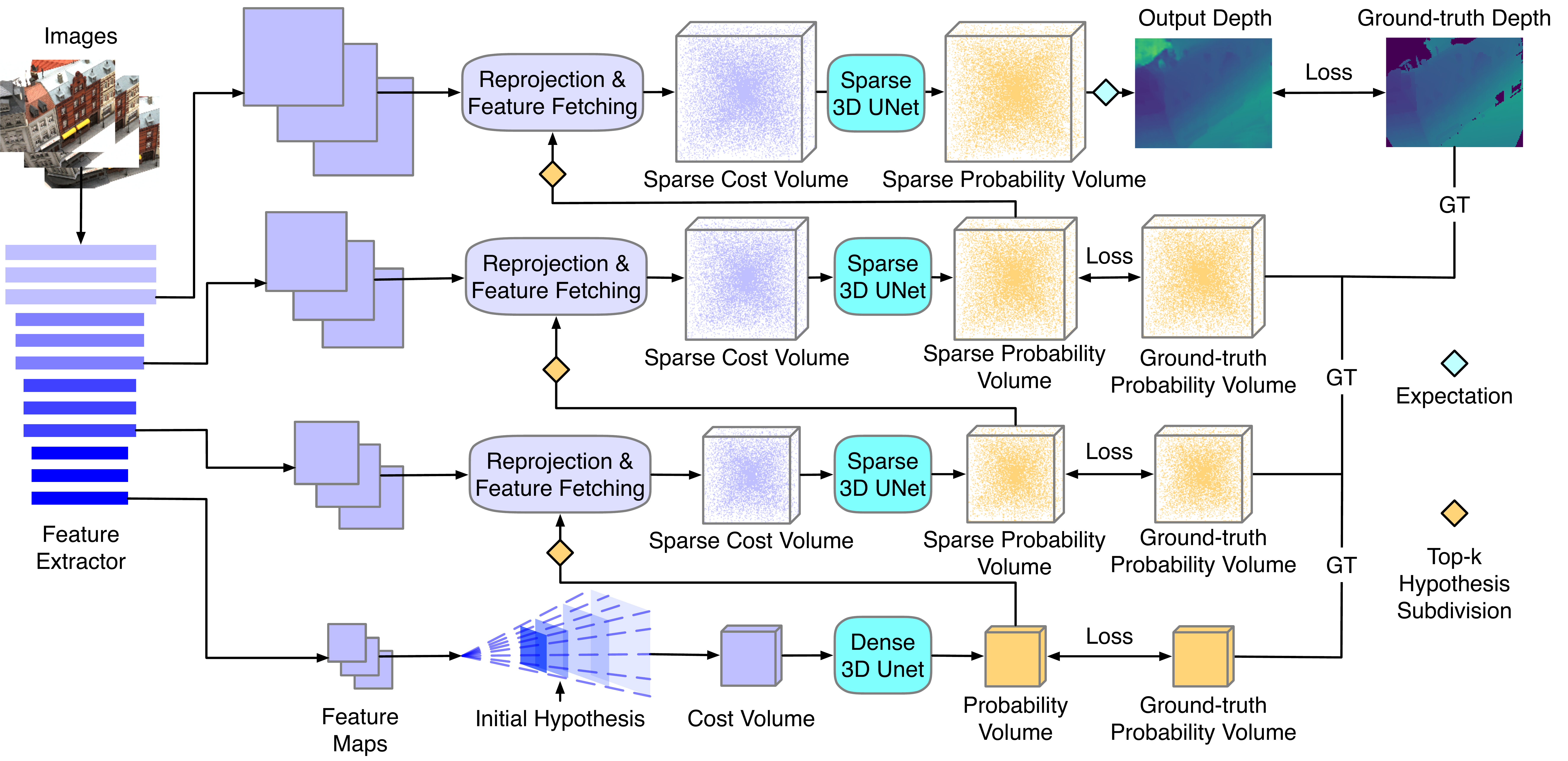}
	\caption{Network Structure. We first build the feature pyramid from the source and reference images. We then build the cost volume pyramid based on the modelling of pixel-wise non-parametric depth probability distribution. Specifically, the cost volume at each level is built based on the depth samples of top K probability from the previous level. The cost volume is sparse and aggregated by sparse convolution. The depth map ${\bf D}^0$ is inferred at the full resolution level. }
	\label{fig:networkStructure}
	\end{center}
	\vspace{-0.7cm}
\end{figure*}
Let ${\bf I}_0\in \mathbb{R}^{H\times
W\times 3}$ be the reference image, where $H$ and $W$ define its dimensions; and let $\{{\bf I}_i\}_{i=1}^N$ be its $N$ neighboring source images.~Assume $\{{\bf K}_i, {\bf R}_i, {\bf
t}_i\}_{i=0}^N$ are the corresponding camera intrinsic parameters, rotation
matrix, and translation vector for all views. Our goal is to infer the depth map $\mathbf{D}\in \mathbb{R}^{H\times W}$ for ${\bf I}_0$ from $\{{\bf I}_i\}_{i=0}^N$. 

Figure~\ref{fig:networkStructure} shows the overall pipeline of our approach to performing depth inference based on a cost volume pyramid built in a coarse-to-fine manner. Unlike existing works, the key idea of our approach is to build the cascaded cost-volume based on the local search around the top K depth hypotheses, which we obtain by modeling the pixel-wise depth probability distribution. Below, we first introduce our the non-parametric depth probability distribution modeling; then the cost volume pyramid built using those non-parametric distributions and the depth map inference process. Finally, we provide details of the loss function.

\subsection{Non-parametric depth distribution modelling}
Existing methods assume the depth $d$ of a pixel $p$ follows a unimodal probability distribution ${\bf P}_{p}(d)$. Under this assumption, the estimated depth $ \hat{d}(p)$ is usually defined as the expectation of this distribution, approximated as the integral of the product of depth hypothesis $\{d_m\}_{m=1}^{M}$ and its estimated probability along the ray: ${\hat{d}}({p}) = \mathbb{E}[{\bf P}_{{p}}(d)] \approx \sum_{d\in \{d_m\}_{m=1}^{M}}d{\bf P}_{p}(d)$.

\begin{figure}[!t]
	\begin{center}
\includegraphics[width=0.85\linewidth]{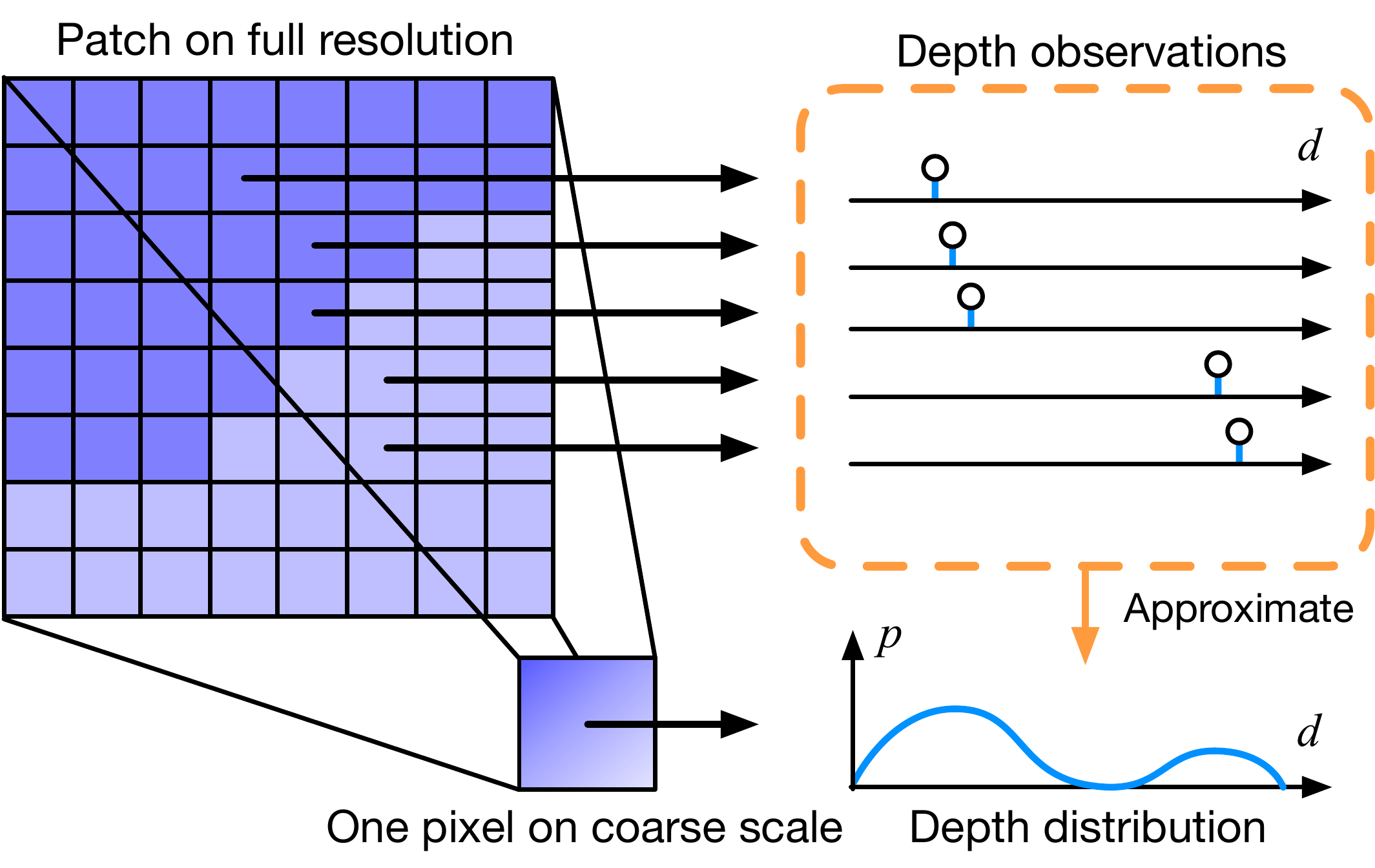}
    \vspace{-0.2cm}
	\caption{The depth distribution of a coarse pixel can be approximated by the depth observations in the corresponding patch on full resolution depth map}
	\label{fig:patchdepth}
	\end{center}
	\vspace{-0.8cm}
\end{figure}

The unimodal depth distribution is a valid assumption if the discrete depth map is of sufficiently high resolution and can approximate the continuous depth distribution well~\cite{tosi2021smd}. However, pixels at lower resolutions could be the projection of a group of 3D points with different depth values, especially for 3D structures across the boundaries of an object with depth discontinuities, which, as shown in  Fig.~\ref{fig:patchdepth}, are inherent of a multi-modal distribution.

\begin{figure*}
	\begin{center}
    \includegraphics[width=\linewidth]{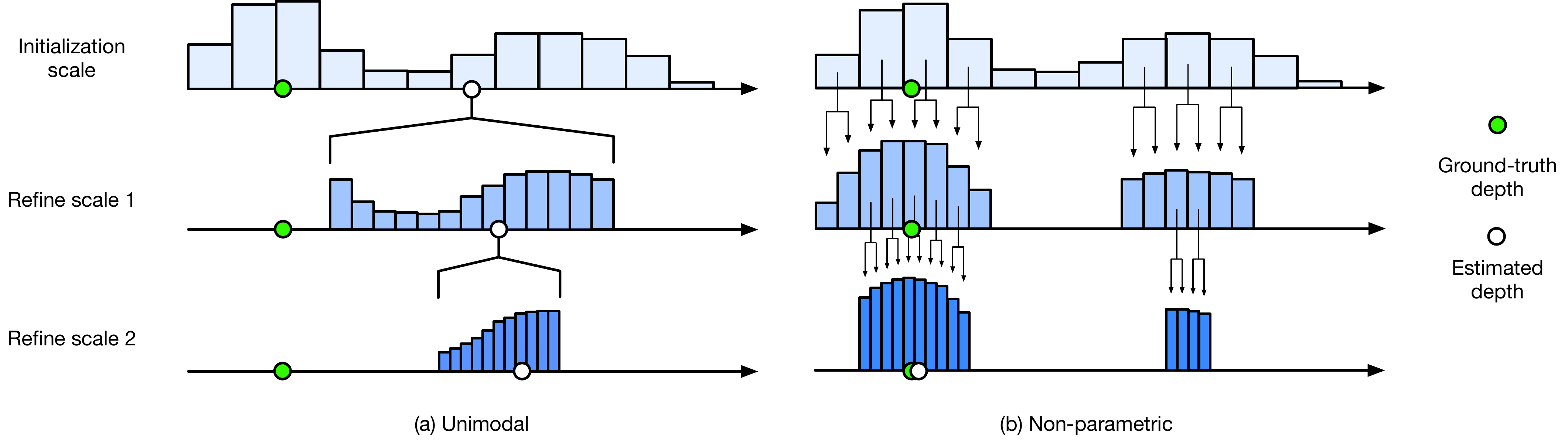}
	\caption{Unimodal and non-parametric depth searching. (a) Existing method assume unimodal distribution that can cause incorrect depth estimation. (b) Our non-parametric depth modeling can estimate correct depth from multi-modal depth distribution.}
	\label{fig:uni_vs_nonp}
	\end{center}
	\vspace{-0.7cm}
\end{figure*}

As shown in Fig.~\ref{fig:uni_vs_nonp}a,  existing cascaded cost volume-based works using a unimodal distribution to represent those pixels might result in incorrect depth estimation. The estimated depth, defined as the expectation of the distribution, might be too far away from any of the depth modes; therefore, it would not be possible to recover it
during the following refinement steps. That is, the algorithm makes an inaccurate early decision, the error of which will be propagated to subsequent modules.  Instead, we introduce a non-parametric depth probability modeling for each pixel to deal with pixels with arbitrary distributions. Specifically, given a pixel $p$ at the coarse level $l$, its depth $d_p$ follows a continuous probability distribution. We approximate this continuous distribution $\mathbf{P}^l({d_{{p}}})$ on a set of discrete depth hypotheses (discrete samples) $\{d^{{l}}_{p,m}\}_{m=1}^{M^l}$.


Next, we introduce our framework for depth inference based on this non-parametric depth probability distribution modeling.

\subsection{Cost volume pyramid}

We construct the cost volume pyramid for depth inference using a feature pyramid to extract features $\mathbf{f}^l$, $l \in \{0...L\}$, where $l=L$ refers to the coarsest level in the smallest resolution and $l=0$ refers to the finest level corresponding to the full resolution. 

\noindent\textbf{Regular cost volume for depth initialization.}
We first build the cost volume at level $L$.~Assume $\{\mathbf{f}^L_i\}_{i=0}^N$ refers to the feature maps extracted from the reference view and the $N$ source views. 

Given a pre-defined global depth searching range $[d_{min},d_{max}]$, we uniformly sample $\{d_m\}_{m=1}^{M^L}$ depth values corresponding to $M^L$ fronto-parallel planes on the \textit{inverse} depth space~\cite{xu2020pvsnet}. Here $d_{min}$ and $d_{max}$ represent the lower and upper bound of the searching range, respectively. Note that a sampled depth $d\in \{d_m\}_{m=1}^{M^L}$ represents a plane parallel to the image plane of the reference camera. We use the differential homography~\cite{yao2018mvsnet} computed for the plane defined by the depth $d$ to warp features of the source view $i$ to the reference view $\mathbf{f}^L_{i\rightarrow 0,d}$. We compute the matching cost as the Group-wise correlation of $G^L$ groups between the reference feature $\mathbf{f}^L_{0}$ and warped source feature $\mathbf{f}^L_{i\rightarrow 0,d}$. Then, we estimate the cost maps for each depth hypothesis plane and concatenate them into a cost volume $C^L \in \mathbb{R}^{H^L\times W^L\times M^L\times G^L}$. Here $W^L$, and $H^L$ denote the image size at level $L$. We adopt a view aggregation module similar to~\cite{xu2020pvsnet} to estimate a visibility map and fuse matching cost computed from different source views.

\begin{figure}[!t]
	\begin{center}
\includegraphics[width=0.95\linewidth]{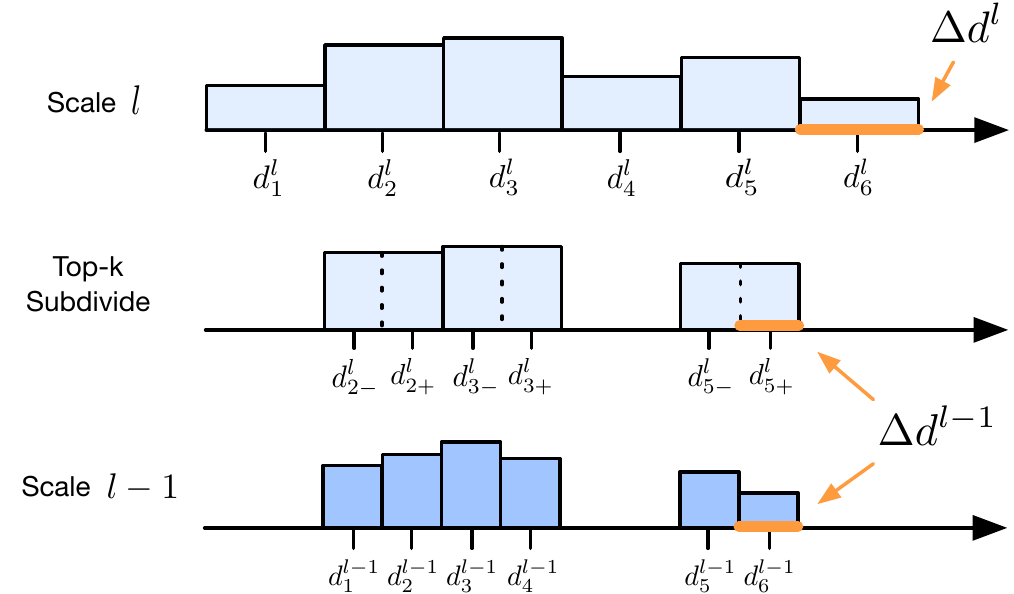}
    \vspace{-0.2cm}
	\caption{We generate new hypothesis by selecting top-k hypothesis from coarse level and equally subdivide them along the ray.}
	\label{fig:subdivide}
	\end{center}
	\vspace{-0.8cm}
\end{figure}

Given this regular cost volume $C^L$, we use a regular 3D-UNet similar to \cite{yang2020cvp} for cost aggregation. 
The output of this initial cost aggregation network is a probability volume denoted as $P^L\in \mathbb{R}^{H^L\times W^L\times M^L}$ which defines the non-parametric depth probability distribution for each pixel, represented by the probability for each depth sample.

Then, we explore the pixel-wise depth samples with the top $K$ probabilities to build the cost volume for the next level.
\begin{figure*}
	\begin{center}
    \includegraphics[width=0.9\linewidth]{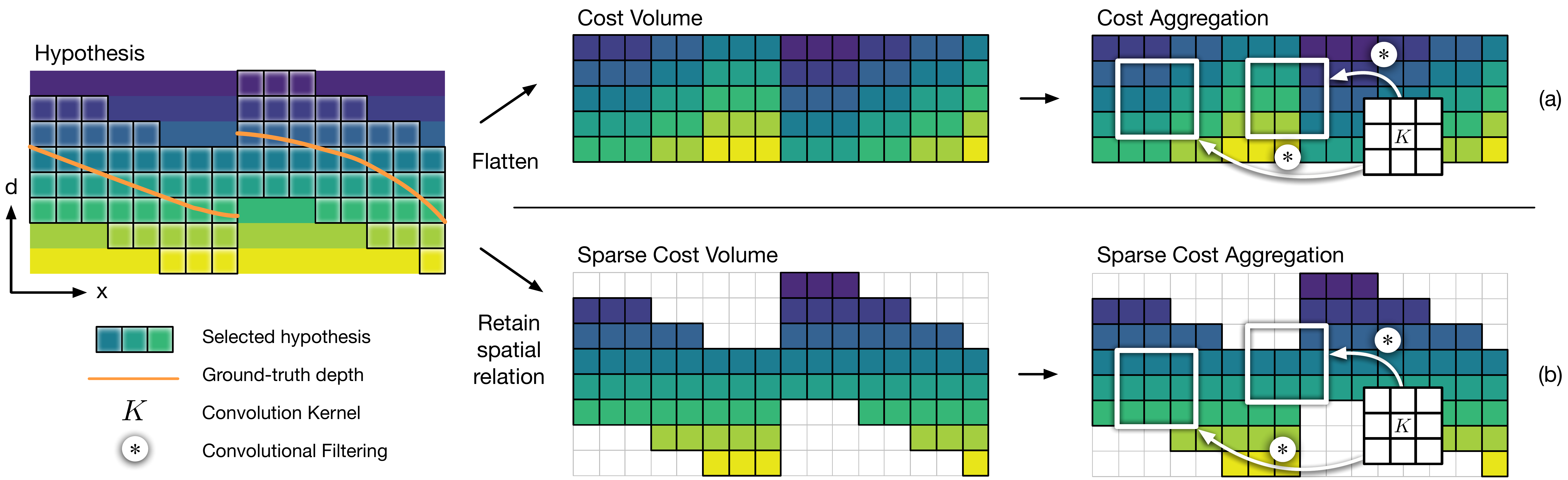}
	\caption{Sparse cost volume and sparse cost aggregation. Color indicates relative depth of hypothesis. (a) Existing methods build flattened cost volume that have spatial ambiguity. (b) We build sparse cost volume and use sparse cost aggregation to retain rigid spatial relation.}
	\label{fig:sparse}
	\end{center}
	\vspace{-0.7cm}
\end{figure*}

\noindent\textbf{Sparse Cost volume for depth refinement.}
 Without loss of generality we ignore the pixel index from now on.
Let $\{d^{l}_{Q_i}\}_{i=1}^K$ define the depth samples of top K estimated probability at level $l$, where $\{Q_i\}_{i=1}^K$ define the top K indices and $\Delta d^{l}$ is the corresponding depth 
search interval. To perform local search around the K possible depth samples obtained from level $l$, we define the depth samples for level $l-1$ by subdividing each selected depth sample at level $l$ into two samples, see Fig.~\ref{fig:subdivide}. This process is formulated for each pixel as 
\begin{equation}
\begin{aligned}
    d^l_{m-} &= d^l_m - \tfrac{1}{4}\Delta d^l \\
    d^l_{m+} &= d^l_m + \tfrac{1}{4}\Delta d^l,
\end{aligned}
\end{equation}
where $m\in \{Q_i\}_{i=1}^K$. The depth samples for level $l-1$ are estimated as ${\bf S}^{l-1} =\{d^l_{m-},d^l_{m+}|m\in \{Q_i\}_{i=1}^K \}$ with the updated depth searching interval $\Delta d^{l-1} = 0.5 \Delta d^{l}$. 

Due to the resolution difference across levels,
${\bf S}^{l-1}$ is shared by pixels within a patch at level $l-1$ that corresponds to a pixel at level $l$.
Given the pixel-wise depth samples ${\bf S}^{l-1}$, we then build the cost volume
to model the depth probability distributions for pixels at level $l-1$. As depth
samples are formed in a pixel-wise manner, the relative spatial locations 
among neighboring 3D points are not reserved. We therefore leverage the 
sparse cost volume structure and aggregate information based on sparse convolutions. To this end, we define the sparse cost volume as $C^{l-1} =\{(p_k,c_k)\}_{k=1}^{K^{l-1}}$,
where $c_k$ defines the matching cost, $p_k = (x_k, y_k, z_k)$ defines the 
3D coordinate computed from a depth sample, $K^{l-1}$ defines the total
number of 3D points converted from depth samples. Given the camera projection matrix,
we project $p_k$ to other source views and the cost is similarly calculated as that
for pixels at level $L$ as described in the previous section.

\noindent \textbf{Sparse cost aggregation network.} As the sparse cost volume can not be efficiently aggregated by regular dense 3D convolutions, we build a sparse cost aggregation network to aggregate cost utilizing the rigid spatial relation stored in $p_k$. Specifically, the basic block of our network consists of three layers of sparse 3D convolution with factorized kernels on each dimension, a sparse batch normalization layer and a sparse ReLU activation~\cite{szegedy2016rethinking}. We provide the detailed structure of the network in the supplementary material. The output of the sparse cost aggregation network is the probability distribution ${\bf P}^{l-1}_{{p}}(d)$, which is used as input to build the cost volume in the next refinement level, see Fig.~\ref{fig:sparse} (b).

\subsection{Depth Inference at full resolution}
Unlike existing cascade cost volume-based approaches, we only perform the depth inference at the full resolution level $0$. At this level, we approximate the depth of each pixel as the expectation of the estimated distribution,
\begin{equation}
 \hat{d}(p) = \mathbb{E}({\bf P}^0_{p}(d)) = \sum_{m=1}^{M^0}d_m^0{\bf P}^0_p(d^0_m).
\label{eq:output_depth}
\end{equation}

\subsection{Loss Function}
We train the network in a supervised manner. We use the depth probability distribution approximated by the high-resolution depth map observations as ground-truth. In particular, for each pixel, ${p}$, the ground-truth probability distribution $\mathbf{P}^{l}_{gt,{ p}}$ is approximated by the histogram of depth observations from the corresponding patch $\Phi_{p}$ at the full-resolution, normalized by the sum of observations, 
\begin{equation}
\centering
\begin{aligned}
\mathbf{P}^{l}_{gt,p}(d^l_m) &=  \frac{\text{\textit{Hist}}^l_{{p}}(d^l_m)}{\sum_{m=1 }^{m=M^l}\text{\textit{Hist}}^l_{{p}}(d^l_m)}, \\
    \text{\textit{Hist}}^l_{p}(d^l_m) &= \sum_{{p}'\in\Phi_{p}}\begin{cases}
    1 - \frac{|d_{{p'}} - d_m^l|}{\Delta d^l},                & \text{if } |d_{p'} - d^l_m| \leq \Delta d\\
    0,              & \text{if } |d_{p'} - d^l_m| > \Delta d,
\end{cases}
\end{aligned}
\label{eq:gt_prob}
\end{equation}

\noindent where $d_{{p}'}$ is the ground-truth depth value of pixel ${p}'$ in patch $\Phi_{p}$, $d^l_m$ is the depth hypothesis and $\Delta d^l$ is the interval between depth hypotheses. 

Finally, for each hypothesis $d^l_m$ of pixel ${p}$, we compute the loss as the binary cross entropy between the estimated probability $\mathbf{P}^l_{{p}}(d^l_m)$ and ground-truth probability $\mathbf{P}^l_{gt,{p}}(d^l_m)$,
\begin{equation}
\mathcal{L}^l_{p}(d^l_m) = \text{\textit{BCE}}(\mathbf{P}^l_{p}(d^l_m),\mathbf{P}^l_{gt,p}(d^l_m)).
\label{eq:loss}
\end{equation}
Empirically, we observe that the ground-truth probability distribution usually concentrates on a few hypotheses, causing an imbalanced number of samples with zero and with non-zero probability. We address this problem balancing the loss as,  
\begin{equation}
\mathcal{L}^l = \sum_{p\in {\Omega}^l} \sum_{m=1}^{M^l}\lambda^l_{{p}}(d^l_m) \mathcal{L}^l_{{p}}(d^l_m),
\label{eq:sum_loss}
\end{equation}
\begin{equation}
\lambda^l_{p}(d^l_m) = \begin{cases}
    1-\sigma^l,                & \text{if } \mathbf{P}^l_{gt,p}(d^l_m)>0\\
    \sigma^l,              & \text{if } \mathbf{P}^l_{gt,p}(d^l_m)=0
\end{cases}
\label{eq:class_weight1}
\end{equation}
\begin{equation}
\sigma^l = \frac{
    \sum_{m=1}^{M^l}\sum_{{p}\in \Omega^l}\mathbbm{1}(\mathbf{P}^l_{gt,p}(d^l_m)>0)
    }{H^l \times W^l \times M^l},
    \label{eq:class_weight2}
\end{equation}

\noindent where $\Omega^l$ defines the image coordinate domain at level $l$, $\sigma^l$ represents the percentage of hypothesis with probability larger than zero. At the final level, we supervise the depth estimation by the $l_1$ norm measuring the distance between ground-truth depth map and final estimated depth map.
\begin{equation}
    \mathcal{L}^0 = \sum_{p\in\Omega^0}||{\bf D}_{gt}^0(p) - {\bf D}^0(p)||_1.
\end{equation}\label{eq:l1_loss}
The overall loss is a weighted sum of the BCE loss on coarse scales and the $l_1$ loss on the final level, 
\begin{equation}
    \mathcal{L} = \sum_{l=0}^{L} w^l \mathcal{L}^l
\end{equation}

\noindent where $w^l$ is the weight for the loss at level $l$.

\section{Experiments}
In this section, we demonstrate the performance of our
approach with a comprehensive set of experiments in
standard benchmarks. Below, we first describe the 
datasets and benchmarks, the implementation details, and then, present and analyze our results.

\begin{figure}[!t]
	\begin{center}
\includegraphics[width=0.95\linewidth]{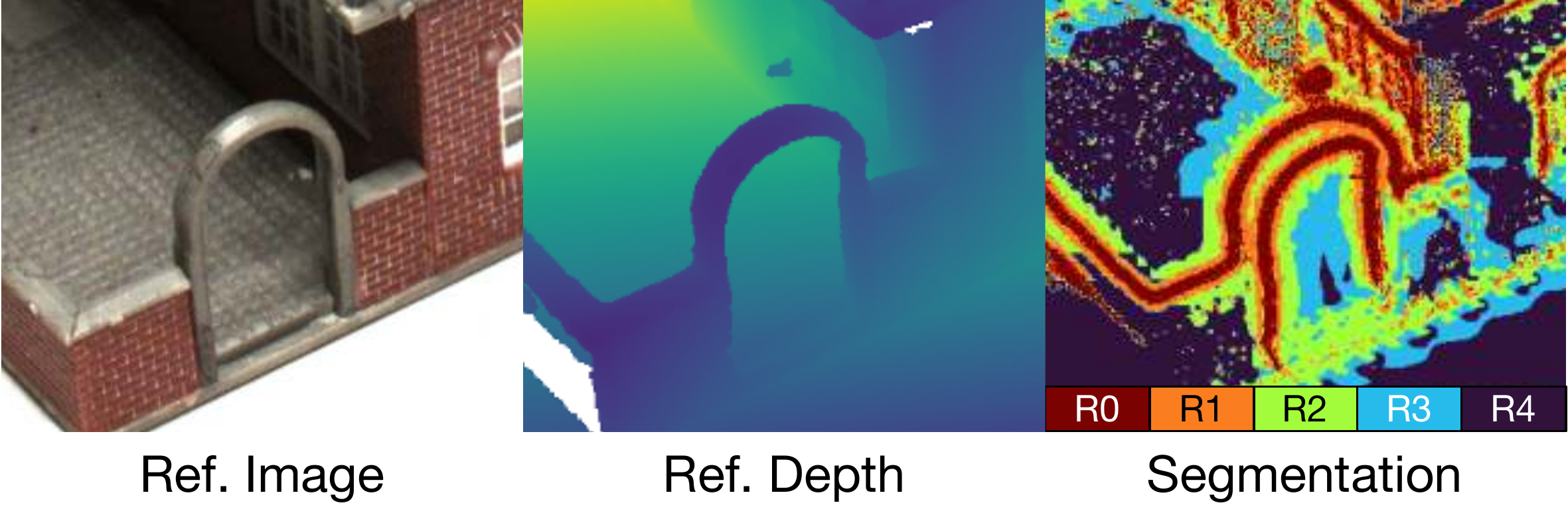}
    \vspace{-0.3cm}
	\caption{We segment ground-truth depth map into five regions corresponding to different depth smoothness and evaluate depth estimation accuracy on each region.}
	\label{fig:seg}
	\end{center}
	\vspace{-0.9cm}
\end{figure}

\begin{figure*}[t]
\begin{center}
\setlength\tabcolsep{1pt}
\vspace{-0.55cm}
\small
\begin{tabular}{cccccc}
\multirow{3}{*}[1.55cm]{\includegraphics[width=0.29\linewidth]{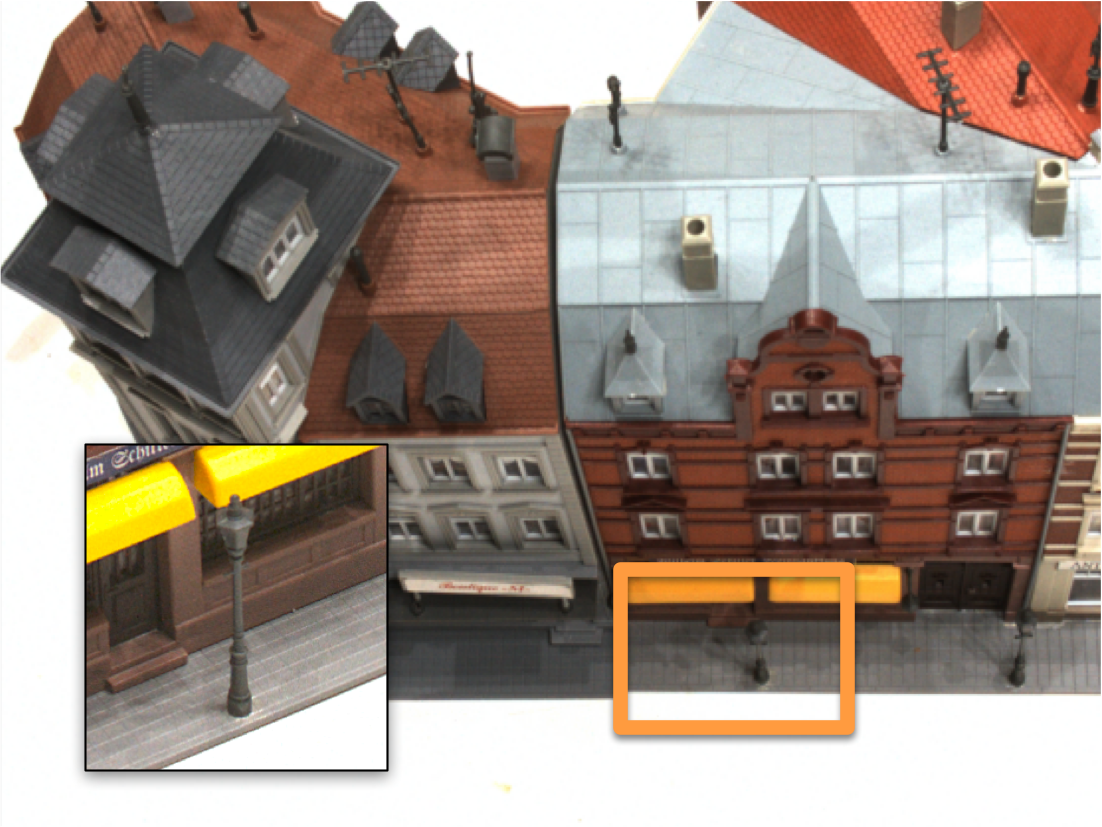}} 
&  \includegraphics[width=0.15\linewidth]{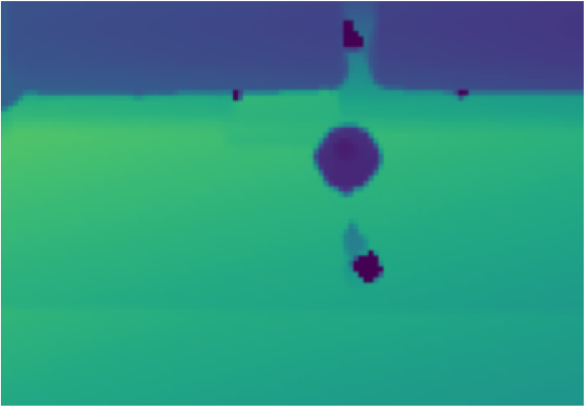}
&  \includegraphics[width=0.15\linewidth]{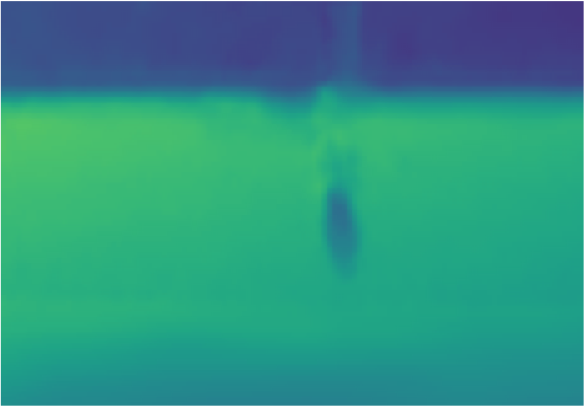}
&  \includegraphics[width=0.15\linewidth]{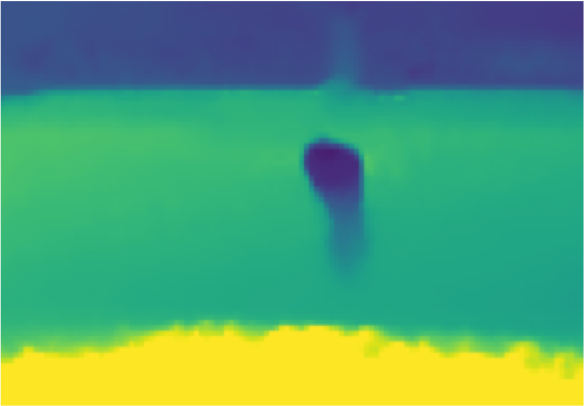}
&  \includegraphics[width=0.15\linewidth]{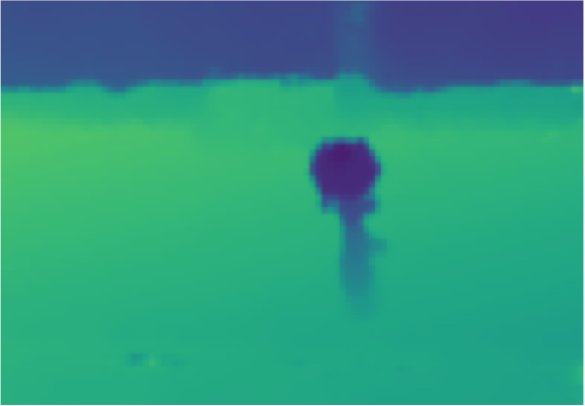}
&  \includegraphics[width=0.032\linewidth]{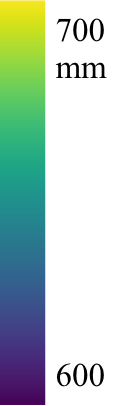} \\  

&  \includegraphics[width=0.15\linewidth]{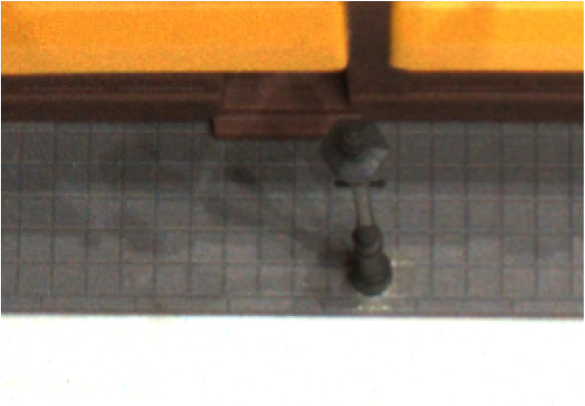}
&  \includegraphics[width=0.15\linewidth]{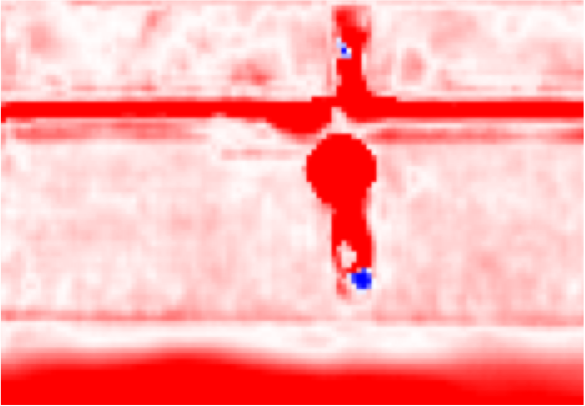}
&  \includegraphics[width=0.15\linewidth]{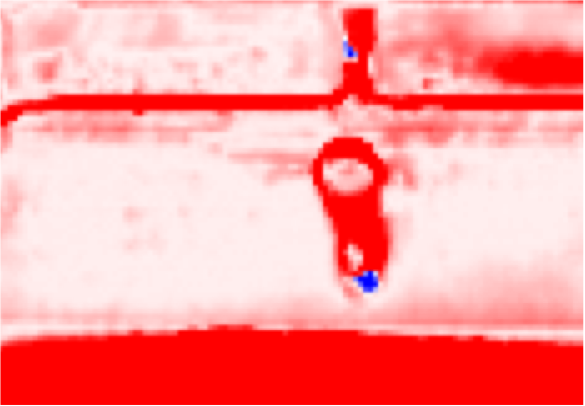}
&  \includegraphics[width=0.15\linewidth]{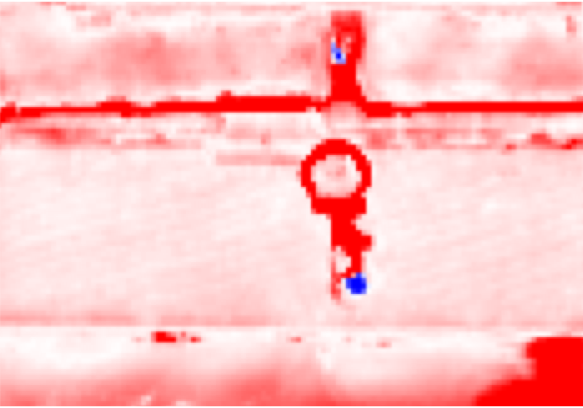}
&  \includegraphics[width=0.032\linewidth]{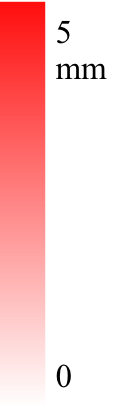} \\ 

\multirow{3}{*}[1.55cm]{\includegraphics[width=0.29\linewidth]{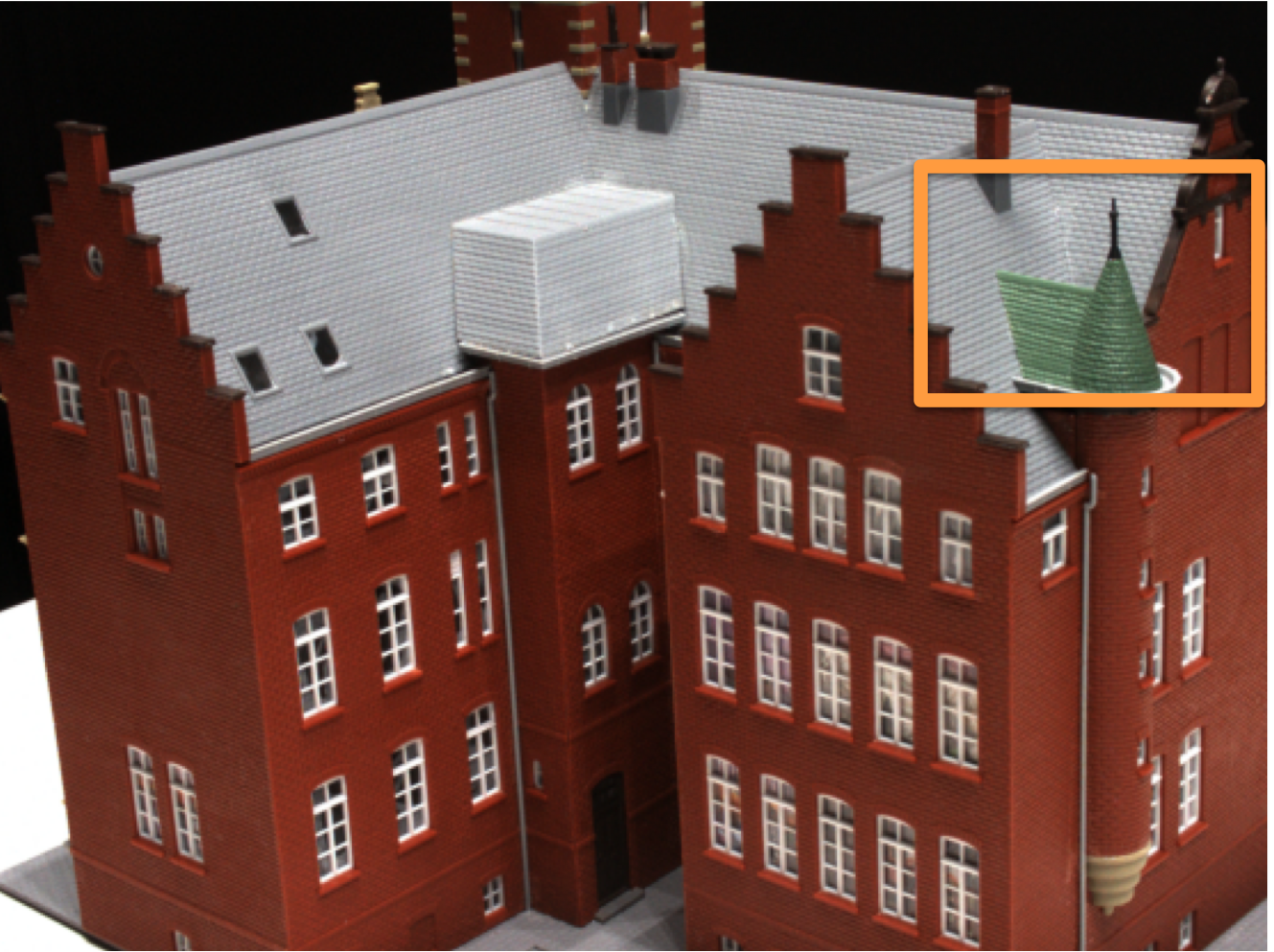}} 
&  \includegraphics[width=0.15\linewidth]{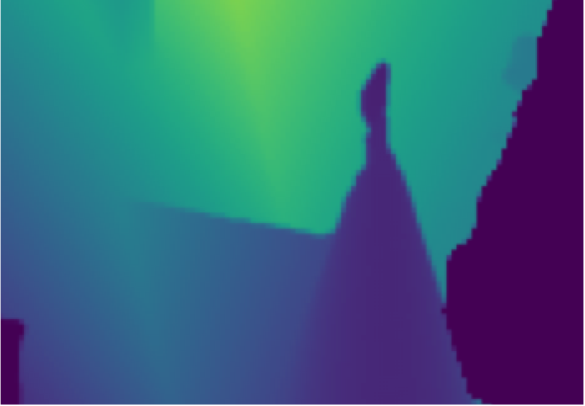}
&  \includegraphics[width=0.15\linewidth]{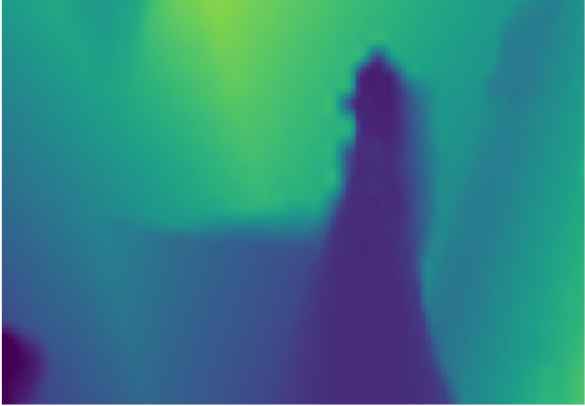}
&  \includegraphics[width=0.15\linewidth]{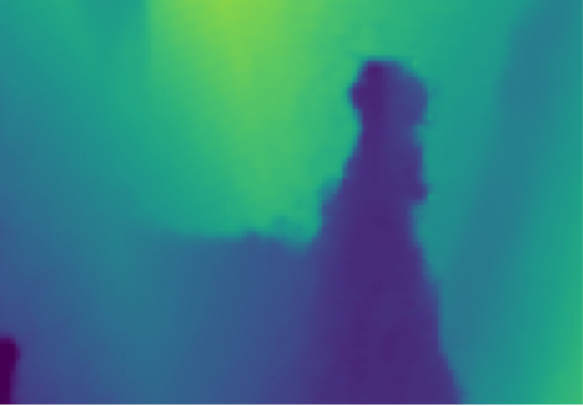}
&  \includegraphics[width=0.15\linewidth]{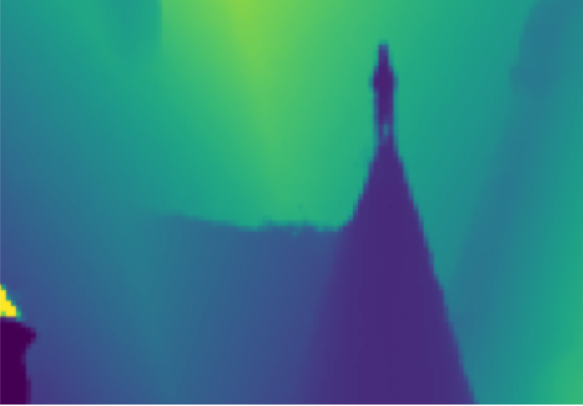}
&  \includegraphics[width=0.032\linewidth]{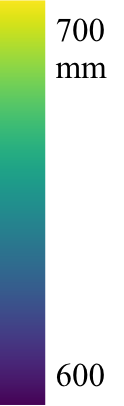} \\  

&  \includegraphics[width=0.15\linewidth]{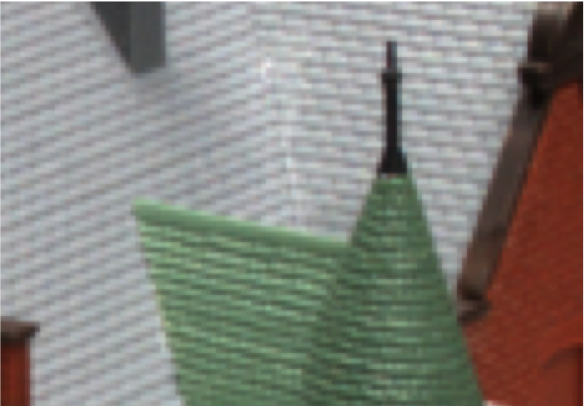}
&  \includegraphics[width=0.15\linewidth]{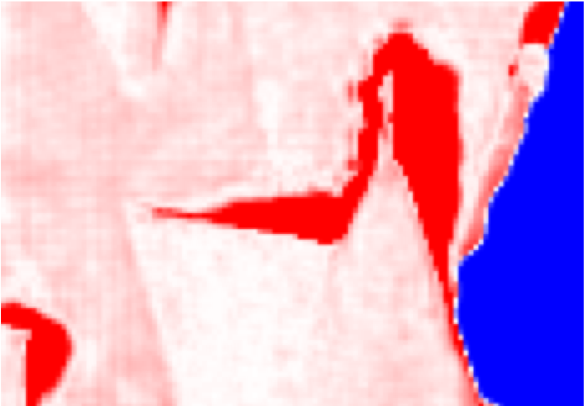}
&  \includegraphics[width=0.15\linewidth]{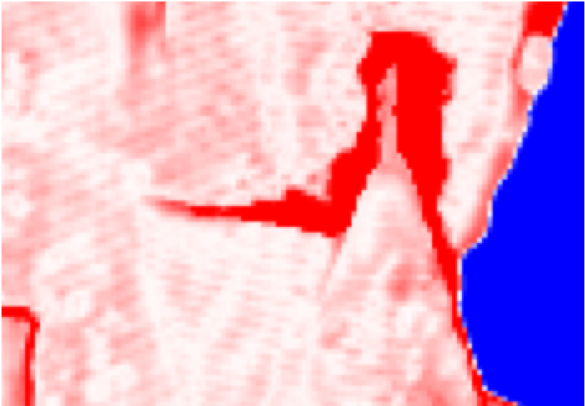}
&  \includegraphics[width=0.15\linewidth]{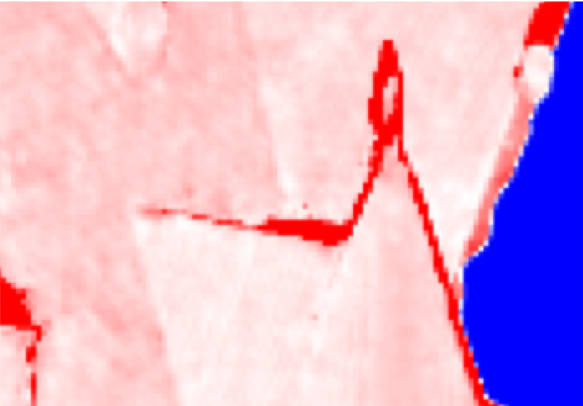}
&  \includegraphics[width=0.032\linewidth]{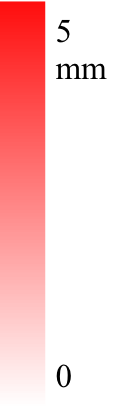} \\ 
Ref. Image
&  Ground-truth
&  Unimodal baseline
&  PatchmatchNet \cite{wang2020patchmatchnet}
&  Ours
&                   \\         
\end{tabular}
\end{center}
\vspace{-0.6cm}
\captionof{figure}{\textbf{DTU dataset.} Example qualitative results on small objects and depth boundaries. Every upper row shows the ground-truth and estimated depth maps and under row shows the estimation error comparing with ground-truth depth map. Areas with no ground-truth are marked blue. Our method based on non-parametric depth distribution modeling is more accurate on small objects and boundary regions. }\label{fig:boundary2}
\vspace{-0.3cm}
\end{figure*}

\subsection{Datasets}\label{sec:dataset}
\noindent\textbf{DTU dataset}\cite{aanaes2016large} is a large-scale MVS dataset providing 124 scenes captured from 64 views under 7 lighting conditions. 3D points captured by structured-light scanners are provided for each scene as reference reconstruction. The dataset provides a color image and the corresponding camera parameters for each view. 

\noindent\textbf{BlendedMVS dataset}\cite{yao2020blendedmvs} is a large-scale synthetic MVS dataset. It contains more than 17000 MVS training samples covering a variety of 113 scenes, including architectures, sculptures and small objects.

\noindent\textbf{Tanks and Temples}\cite{knapitsch2017tanks} dataset provides both indoor and outdoor scenes under real-life lightning conditions with large scale variations. We evaluate the generalization ability of our model on this public benchmark dataset.

\subsection{Implementation}\label{sec:implementation}
We implement the proposed model using \textit{pytorch}\cite{pytorch} and use the \textit{Torchsparse}\cite{tang2020searching} library for the sparse cost volume and the sparse cost aggregation. 

\noindent\textbf{Training}
We train our model on the DTU dataset using downsampled and cropped images of size $640\times 512$ and their corresponding depth map. This map is rendered from a surface mesh generated by Screened Poisson Surface Reconstruction of the provided reference point cloud. We use the same split of training, validation and test set as defined in~\cite{yao2018mvsnet}. We set the number of levels to $(L+1)=4$ and the number of hypothesis to $\{M^l\}_{l=0}^{L} = \{8,16,32,48\}$. We use 5 views for training and adopt the similar anti-noise training strategy as in~\cite{wang2020patchmatchnet}. We use Adam optimizer with $0.0005$ learning rate and a batchsize of 2 on a Nvidia RTX Titan GPU. To demonstrate the generalization ability of our approach, we train our model on BlendedMVS \cite{yao2020blendedmvs} training set and test on Tanks and Temples dataset without any fine-tuning.

\noindent\textbf{Metrics}
We follow the evaluation protocol used by \cite{yao2018mvsnet,yao2019recurrent,yang2020cvp}. On the DTU dataset, we report the~\emph{accuracy},~\emph{completeness} and~\emph{overall score} of the reconstructed point clouds. \emph{Accuracy} measures the distance from the estimated point clouds to the ground truth in millimeters and~\emph{completeness} measures the distance from ground truth point clouds to the estimated ones~\cite{aanaes2016large}. The~\emph{overall score} is the average of accuracy and completeness~\cite{yao2018mvsnet}. On Tanks and Temples, we report the \textit{f-score} for each scene and the average of all scenes. 

We also evaluate the depth estimation accuracy specifically on depth boundaries. To this end, we use a Laplacian pyramid simulating a band-pass filter to segment the ground-truth depth map into non-overlapping regions of different depth smoothness, see Fig. \ref{fig:seg}. We first build a depth map pyramid $\{{\bf D}^q_{gt}\}_{q=0}^Q$ of $Q+1$ levels by repeatedly downsampling the ground-truth depth map ${\bf D}_{gt}^0$. We then build the Laplacian pyramid of depth by taking the difference between adjacent levels in $\{{\bf D}^q_{gt}\}_{q=0}^Q$, where the lower level depth is upsampled to match the size of the upper one. 

Each level of the Laplacian pyramid contains depth structures present at a particular scale, which we use as a mask to segment the depth map into different smoothness regions. In particular, we use 5 levels to segment the depth map into five regions $\{R0, R1, R2, R3, R4\}$, see Fig. \ref{fig:seg}. The region $R0$ corresponds to the highest level of the Laplacian pyramid and contains pixels with the most abrupt depth changes usually caused by depth boundaries or small objects. The lower levels correspond to regions with intermediate and smooth depths. We report the average depth error for each of the five region as the average $l_1$ distance in millimeter between estimated depth and ground-truth depth.

\noindent\textbf{Evaluation}
We set the number of hypothesis per level as $\{M^l\}_{l=0}^{L} = \{8,16,32,96\}$ for testing. On the DTU dataset, we test on the full-resolution images of size $1600\times 1184$ and set number of views to $N=5$ with \cite{galliani2016gipuma} for fusion. On the Tanks and Temples dataset, we use images of size $1920\times 1080$ with $N=11$ views, camera parameters generated by Colmap\cite{schoenberger2016mvs,schoenberger2016sfm} and the fusion method in~\cite{yan2020d2hc}.

\begin{table}[!t]
\begin{center}
\small
\begin{tabular}{l|c|ccc}
\hline
Methods                             & Views & Acc.   & Comp.  & Overall \\ \hline\hline
D2HC-RMVSNet\cite{yan2020d2hc}      & 7                & 0.395 & 0.378 & 0.386   \\ \hline
Vis-MVSNet\cite{zhang2020vismvsnet} & 5                & 0.369 & 0.361 & 0.365   \\ \hline
PVA-MVSNet\cite{zhang2020vismvsnet} & 7                & 0.379 & 0.336 & 0.357   \\ \hline
AA-RMVSNet\cite{wei2021aarmvsnet}   & 7                & 0.376 & 0.339 & 0.357   \\ \hline
CasMVSNet\cite{gu2020cascade}   & 5                & 0.325 & 0.385 & 0.355   \\ \hline
EPP-MVSNet\cite{ma2021epp}          & 5                & 0.413 & 0.296 & 0.355   \\ \hline
PatchmatchNet\cite{wang2020patchmatchnet} & 5                & 0.427 & 0.277 & 0.352   \\ \hline
CVP-MVSNet\cite{yang2020cvp}                & 5                & \textbf{0.296} & 0.406 & 0.351   \\ \hline
UCSNet\cite{cheng2020ucsnet}                & 5                & 0.338 & 0.349 & 0.344   \\ \hline
BP-MVSNet\cite{sormann2020bp} & 5 & 0.333  & 0.320 & 0.327  \\ \hline
PVSNet\cite{xu2020pvsnet}             & 11              & 0.337 & 0.315 & 0.326   \\ \hline
Ours                      & 5                & 0.356 & \textbf{0.275} & \textbf{0.315}   \\ \hline
\end{tabular}
\caption{\textbf{DTU dataset.} Quantitative results of reconstruction quality; lower is better. Our model outperforms all existing methods on mean completeness and overall score.} 
\label{table:dtu}
\end{center}
\vspace{-0.9cm}
\end{table}

\subsection{Results on DTU dataset}\label{sec:dtu}
We first compare the overall reconstruction quality of our proposed method with existing methods on the DTU dataset. As summarized in Tab. \ref{table:dtu}, our method outperforms all existing methods on both \textit{mean completeness} and \textit{overall score}. We also analyze the quality of the reconstruction on the boundaries using the average depth error on the five regions of the Laplacian pyramid. For comparison, we consider PatchmatchNet~\cite{wang2020patchmatchnet} and a recurrent method D2HC-RMVSNet~\cite{yan2020d2hc}. As a baseline, we consider a variation of our method using a unimodal distribution. As summarized in Tab. \ref{table:boundary}, our method outperforms the others on the boundary regions. Qualitative results in Fig. \ref{fig:startfig} and Fig. \ref{fig:boundary2} also demonstrate our method is more accurate on boundary regions and small objects producing sharper depth discontinuities. Efficiency-wise, our model takes 6054 MB GPU memory and 1.2s to estimate a full-resolution depth map, which is on par with existing cost volume-based methods.

\begin{table}[!t]
\small
\begin{tabular}{l|ccccc}
\hline
Method & R0       & R1   & R2       & R3  & R4      \\ \hline\hline
D2HC-RMVSNet\cite{yan2020d2hc}          & 7.20 & 2.33  & 2.59 & 5.14 & 10.2 \\ \hline
PatchmatchNet\cite{wang2020patchmatchnet}         & 9.54  &  2.37 & 2.40  & 4.72 & 9.15  \\ \hline
Unimodal baseline & 8.85  & 2.39 & 2.15 & 3.97 & 9.08  \\ \hline
Ours & \textbf{6.48}  & \textbf{1.94} & \textbf{1.81}   & \textbf{3.85} & \textbf{8.61}  \\ \hline
\end{tabular}
\caption{\textbf{DTU dataset.} Performance on regions of different depth smoothness. Our method can achieve lowest error on boundary region (R0).}
\label{table:boundary}
\vspace{-0.5cm}
\end{table}

\subsection{Results on Tanks and Temples}\label{sec:generalize}
\begin{figure*}[!t]
\begin{center}
\setlength\tabcolsep{1pt}
\vspace{-0.55cm}
\small
\begin{tabular}{cccc}
  \includegraphics[width=0.24\linewidth]{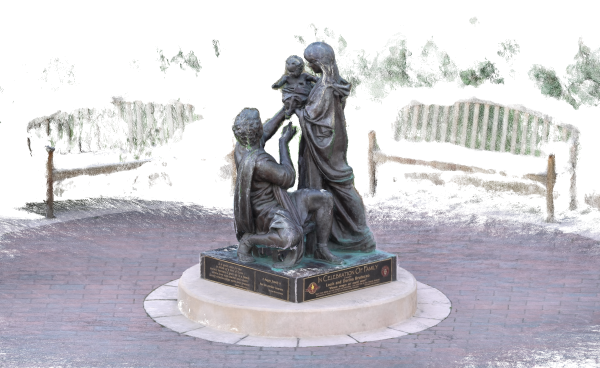}
&  \includegraphics[width=0.24\linewidth]{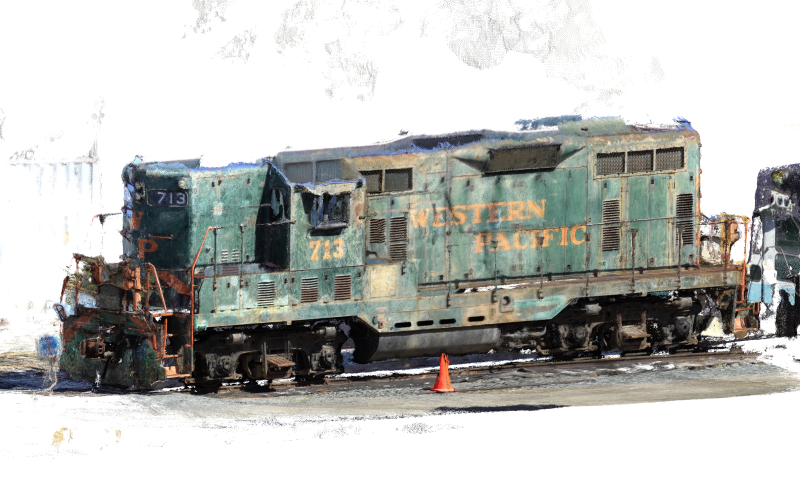}
&  \includegraphics[width=0.24\linewidth]{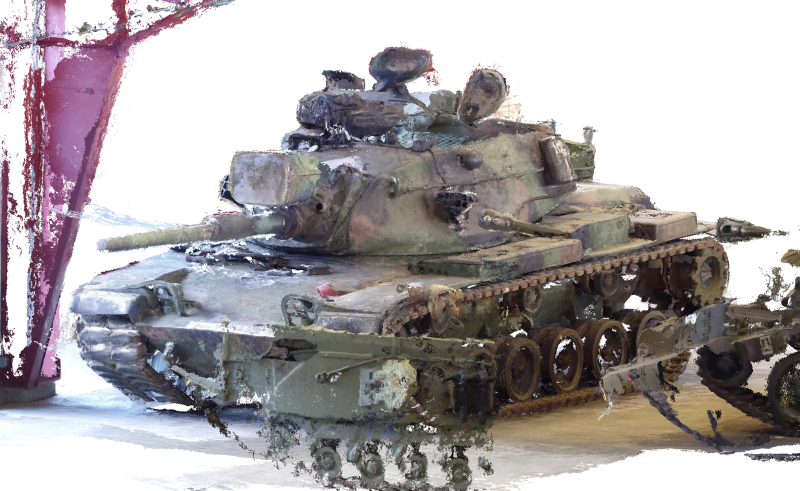}
&  \includegraphics[width=0.24\linewidth]{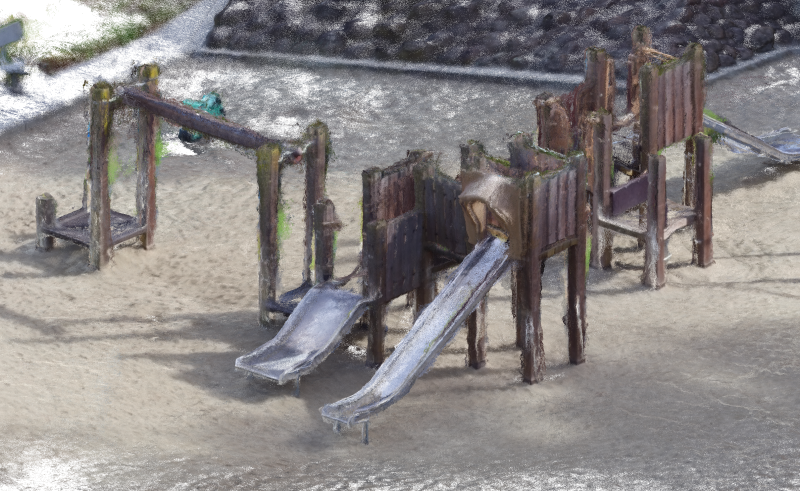} \\
  \includegraphics[width=0.24\linewidth]{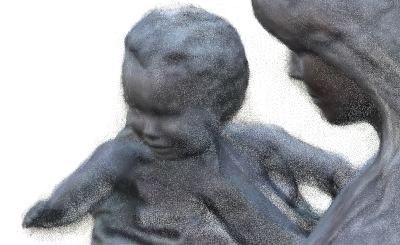}
&  \includegraphics[width=0.24\linewidth]{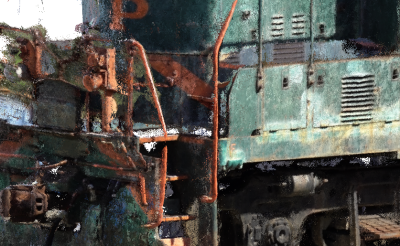}
&  \includegraphics[width=0.24\linewidth]{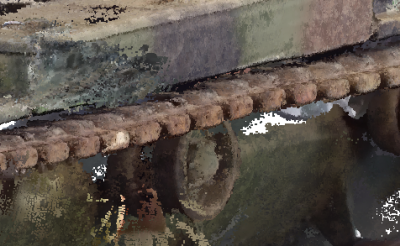}
&  \includegraphics[width=0.24\linewidth]{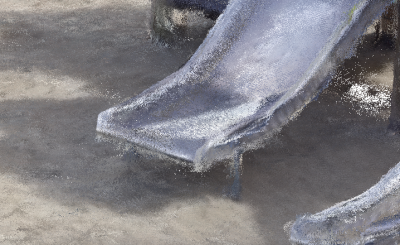} \\     
\end{tabular}
\end{center}
\vspace{-0.5cm}
\captionof{figure}{\textbf{Tanks and Temples dataset.} Example point cloud results. The second row shows details of first row. Best viewed on screen. }\label{fig:tanks}
\vspace{-0.1cm}
\end{figure*}

In this experiment, we evaluate the generalization ability of the proposed method on the Tanks and Temples dataset\cite{knapitsch2017tanks}. The summary of quantitative results is listed in Tab. \ref{table:tanks} and representative qualitative results are shown in Fig. \ref{fig:tanks}. In these results, we can observe that our approach produces sharp and accurate depth estimation on boundary regions and, overall, our model trained on the BlendedMVS dataset achieves competitive performance compared to other recent methods in the benchmark.

\begin{table*}[t]
\begin{center}
\small
\begin{tabular}{l|cccccccccc}
\hline
Methods  & Mean  & Family & Francis & Horse & Lighthouse & M60   & Panther & Playground & Train \\ \hline\hline
AA-RMVSNet\cite{wei2021aarmvsnet}         & 61.51 & 77.77  & 59.53   & 51.53 & 64.02      & 64.05 & 59.47   & 60.85      & 54.90  \\ \hline
EPP-MVSNet\cite{ma2021epp}         & 61.68 & 77.86  & 60.54   & 52.96 & 62.33      & 61.69 & 60.34   & 62.44      & 55.30  \\ \hline
Vis-MVSNet\cite{zhang2020vismvsnet}         & 60.03 & 77.40   & 60.23   & 47.07 & 63.44      & 62.21 & 57.28   & 60.54      & 52.07 \\ \hline
Ours  & 59.64 & 78.93 & 64.09 & 51.82 & 59.42 & 58.39 & 55.71 & 56.07 & 52.71 \\ \hline
D2HC-RMVSNet\cite{yan2020d2hc}       & 59.20  & 74.69  & 56.04   & 49.42 & 60.08      & 59.81 & 59.61   & 60.04      & 53.92 \\ \hline
BP-MVSNet\cite{sormann2020bp}  & 57.60 & 77.31 & 60.90 & 47.89 & 58.26 & 56.00 & 51.54 & 58.47 & 50.41 \\ \hline
PVSNet\cite{xu2020pvsnet}  & 56.88 & 74.00 & 55.17 & 39.85 & 61.37 & 60.22 & 56.87 & 58.02 & 49.51 \\ \hline
PatchmatchNet\cite{wang2020patchmatchnet} & 53.15 & 66.99 & 52.64 & 43.24 & 54.87 & 52.87 & 49.54 & 54.21 & 50.81 \\ \hline
PatchMatch-RL\cite{lee2021patchmatchrl}  & 51.81 & 60.37 & 43.26 & 36.43 & 56.27 & 57.30 & 53.43 & 59.85 & 47.61 \\ \hline
\end{tabular}
\vspace{-0.2cm}
\caption{\textbf{Tanks and Temples dataset.} Quantitative comparison to existing approaches. Our model can achieve competitive performance on Tanks and Temples dataset.} 
\label{table:tanks}
\vspace{-0.45cm}
\end{center}
\end{table*}

\begin{table*}[!t]
\begin{center}
\small
\begin{tabular}{l|cc|cc|ccc}
\hline
\multirow{2}{*}{Method}  & \multicolumn{2}{c|}{Depth distribution}       & \multicolumn{2}{c|}{Cost aggregation} & \multirow{2}{*}{Acc.} & \multirow{2}{*}{Comp.} & \multirow{2}{*}{Overall} \\ \cline{2-5}
                                      & \multicolumn{1}{c|}{Unimodal} & Non-parametric & \multicolumn{1}{c|}{Standard}  & Sparse  &                       &                        &                           \\ \hline\hline
Baseline                           & \multicolumn{1}{c|}{\checkmark}           &            & \multicolumn{1}{c|}{\checkmark}      &         & 0.3498                & 0.3351                 & 0.3425                    \\ \hline
Baseline + non-parametric modeling             & \multicolumn{1}{c|}{}            & \checkmark          & \multicolumn{1}{c|}{\checkmark}      &         & 0.3387                & 0.3792                 & 0.3590                    \\ \hline
Baseline + sparse cost aggregation & \multicolumn{1}{c|}{\checkmark}           &            & \multicolumn{1}{c|}{}       & \checkmark       & 0.3538                & 0.3096                 & 0.3317                    \\ \hline
Ours                                  & \multicolumn{1}{c|}{}            & \checkmark          & \multicolumn{1}{c|}{}       & \checkmark       & 0.3563 & 0.2750 & \textbf{0.3156}                    \\ \hline
\end{tabular}
\vspace{-0.2cm}
\caption{\textbf{DTU dataset.} Quantitative comparison of the contribution of each module in our framework. Using non-parametric depth distribution modeling with standard 3D convolutions can cause performance drop due to spatial ambiguity. Our proposed sparse cost aggregation can improve the performance of unimodal based methods. The proposed non-parametric depth distribution modeling requires the sparse cost aggregation to achieve best reconstruction quality. }\label{table:ablation}
\end{center}
\vspace{-0.80cm}
\end{table*}

\subsection{Ablation Study}\label{sec:ablation}
We provide ablation experiments on the DTU dataset to evaluate the contribution of the proposed modules. We start with a baseline model using standard unimodal distribution and standard 3D convolutions for cost aggregation. Results are shown in Tab. \ref{table:ablation}. Using non-parametric depth distribution modeling with standard 3D convolutions can cause performance drop due to spatial ambiguity. Our proposed sparse cost aggregation can improve the performance of unimodal based methods. The proposed non-parametric depth distribution modeling requires the sparse cost aggregation to achieve best reconstruction quality.

\textbf{Limitations.} Our approach leverages sparse cost volume and sparse convolutions to yield better accuracy. However, sparse convolutions are computationally expensive as they are not fully optimized. This increases the inference time. We plan to improve the efficiency in our future work.

\section{Conclusion}
We proposed an approach for depth inference based on modelling non-parametric depth probability distribution for each pixel.
Our modelling can handle pixels with unimodal and multi-modal depth distributions such as pixels on the boundaries. Our method does not
infer depth at the coarse level which avoids the depth errors made at the early stage and following propagation to the refinement levels. Experimental results show that our approach can achieve superior performance especially for pixels on the boundaries.

\section*{Acknowledgments}
This research is supported by Australian Research Council grants (DE180100628, DP200102274).

{\small
\bibliographystyle{ieee_fullname}
\bibliography{egbib}
}

\end{document}